\documentclass[final, conference]{IEEEtran}
\IEEEoverridecommandlockouts
\usepackage{cite}
\usepackage{amsmath,amssymb,amsfonts}
\usepackage{algorithmic}
\usepackage{graphicx} 
\usepackage{calc} 

\usepackage{xcolor}
\usepackage{relsize} 
\usepackage{cancel} 	
\usepackage{bm}
\usepackage{float}
\usepackage{pgfplots}
\pgfplotsset{width = 10cm, compat = 1.5}
\usepgfplotslibrary{groupplots}
\usepgfplotslibrary{fillbetween}
\usepackage{siunitx}
\def\BibTeX{{\rm B\kern-.05em{\sc i\kern-.025em b}\kern-.08em
    T\kern-.1667em\lower.7ex\hbox{E}\kern-.125emX}}
    
\begin{document}
\title{Robust Optimal Design of Energy Efficient Series Elastic Actuators: Application to a Powered Prosthetic Ankle\thanks{This work was supported by the National Science Foundation under Award Number 1830360 and the National Institute of Child Health \& Human Development of the NIH under Award Number DP2HD080349. The content is solely the responsibility of the authors and does not necessarily represent the official views of the NIH or the NSF. R. D. Gregg holds a Career Award at the Scientific Interface from the Burroughs Wellcome Fund. E. Bol\'ivar, S. Rezazadeh, and R. D. Gregg are with the Departments of Bioengineering and Mechanical Engineering, T. Summers is with the Department of Mechanical Engineering, The University of Texas at Dallas, Richardson, TX, 75080, USA. Email:\{ebolivar, rgregg\}@ieee.org}}

\author{Edgar Bol\'ivar, Siavash Rezazadeh, Tyler Summers, and Robert D. Gregg}

\maketitle

\begin{abstract}
Design of robotic systems that safely and efficiently operate in uncertain operational conditions, such as rehabilitation and physical assistance robots, remains an important challenge in the field. Current methods for the design of energy efficient series elastic actuators use an optimization formulation that typically assumes known operational conditions. This approach could lead to actuators that cannot perform in uncertain environments because elongation, speed, or torque requirements may be beyond actuator specifications when the operation deviates from its nominal conditions. Addressing this gap, we propose a convex optimization formulation to design the stiffness of series elastic actuators to minimize energy consumption and satisfy actuator constraints despite uncertainty due to manufacturing of the spring, unmodeled dynamics, efficiency of the transmission, and the kinematics and kinetics of the load. In our formulation, we express energy consumption as a scalar convex-quadratic function of compliance. In the unconstrained case, this quadratic equation provides an analytical solution to the optimal value of stiffness that minimizes energy consumption for arbitrary periodic reference trajectories. As actuator constraints, we consider peak motor torque, peak motor velocity, limitations due to the speed-torque relationship of DC motors, and peak elongation of the spring. As a simulation case study, we apply our formulation to the robust design of a series elastic actuator for a powered prosthetic ankle. Our simulation results indicate that a small trade-off between energy efficiency and robustness is justified to design actuators that can operate with uncertainty.
\end{abstract}


\section{Introduction}
Series elastic actuators (SEAs) \cite{Pratt1995} have the potential to be a safe, energy efficient, and easy to control actuation scheme for human-robot interaction in rehabilitation and physical assistance robots. SEAs are well suited for force control and impedance control in human-robot interaction, as the elastic element behaves as a soft load cell \cite{Hogan1985,Losey2016}. SEAs encourage safety in robots by potentially reducing the mass of the actuator \cite{Hollander2006} and its associated kinetic energy during impacts. Additionally, SEAs enable elastic collisions for greater safety where impacts might occur \cite{Bicchi2004}. Elastic collisions can also improve energy efficiency in applications subject to periodic impacts such as bipedal locomotion \cite{Hurst2008, Rezazadeh2018}. In addition, compared to rigid actuators, SEAs can reduce energy dissipated by the actuator for periodic tasks \cite{Grioli2015,Bolivar2017}.

The SEA torque and speed bandwidths, reachable virtual and mechanical impedance, stored elastic energy, tolerance to impact loads, peak power output, and energy efficiency \cite{Losey2016, Paine2014} depend on the selection of the SEA's spring stiffness. The requirements of the application ultimately determine which of these criteria should be used for the stiffness design. In this paper, we focus on reducing energy consumption of the SEA, which has the potential to reduce mass and increase battery life of robots for rehabilitation and physical assistance.

Existing methods to design the stiffness that minimizes energy consumption of SEAs, such as natural dynamics \cite{Verstraten2016} and optimization formulations \cite{Rouse2014,Bolivar2017}, assume nominal reference kinematic and kinetic trajectories of the load. These nominal trajectories easily change during operation in human-robot interaction. For example, in the design of an SEA for a powered prosthetic leg, the reference kinematic and kinetic trajectories change as the subject walks with different speeds or wears different accessories, such as a backpack. When the load conditions deviate from their nominal values, the energy consumption and peak power of SEAs may not be optimal \cite{Grimmer2011}. Additionally, the speed and torque requirements for the motor may be outside the motor's specifications, i.e., the task becomes infeasible. For instances when changing stiffness of the SEA makes the task feasible again, a possible solution is to replace SEAs with variable stiffness actuators (VSAs) \cite{Grioli2015}. However, VSAs require additional mechanisms to operate, increasing the mechanical complexity and potentially the mass of the actuator. Thus, we are interested to know if a fixed-stiffness SEA could satisfy the actuator constraints despite uncertainty, and what trade-off a robust design would have with energy consumption.

Acknowledging uncertainty in the reference trajectories leads to more realistic and robust designs. For example, in \cite{Grimmer2011}, the optimal design of series stiffness considered deviation from the nominal trajectories for the application of powered prosthetic legs. The SEA was optimized over both walking and running reference trajectories, but the design did not consider the wide range of other possible tasks and did not analyze the feasibility of the actuator. Brown and Ulsoy \cite{RobertBrown2013} considered task uncertainty by defining the reference trajectory as a sample from a probability distribution of reference trajectories. Their stochastic approach to designing linear parallel elastic elements provided energy savings and constraint satisfaction for 90\% of the reference trajectories, but worst-case scenarios would violate the strict safety requirements of a co-robot. More arbitrary reference motions resulted in stiffer optimal solutions, converging to a rigid actuator for totally arbitrary motion \cite{RobertBrown2013}. However, increasing stiffness may not be the solution when actuator constraints must be satisfied despite uncertainty, as demonstrated in Section \ref{sec:CaseStudy}. Thus, a robust formulation is required to guarantee feasibility of the actuator even in the worst case conditions that could manifest during operation.

\subsection*{Our Contribution}
We present a convex optimization formulation to design the stiffness of an SEA to minimize energy consumption and satisfy actuator constraints despite uncertainty due to manufacturing of the spring, unmodeled dynamics, efficiency of the transmission, and the kinematics and kinetics of the load.  
As actuator constraints we consider: 1) peak motor torque, 2) peak motor velocity, 3) limitations due to the speed-torque relationship of DC motors, and 4) maximum elongation of the spring. We use existing robust optimization techniques to design robust-feasible stiffness values, i.e., satisfy the constraints despite uncertainty \cite{Ben-Tal2009}. 

In general, any optimization problem benefits from a robust feasible solution. However, only a few robust feasible optimization problems, especially convex, can be computationally tractable \cite{Ben-Tal2009}. 
Part of our contribution is to write energy consumption of an SEA's electric motor as a convex-quadratic function of compliance, the inverse of stiffness. This convex-quadratic formulation is not only useful for a robust feasible design, but it also provides an analytical solution for the optimal stiffness in the unconstrained case. Our convex-quadratic formulation allows computation of an optimal value of stiffness within polynomial time \cite{Nesterov94}, which is beneficial for systems that can modify their mechanical stiffness during operation, such as VSAs.

Our formulation applies to any application with periodic motion. We apply our methods to the actuator design of powered prosthetic legs. Battery life and device mass influence the performance of these devices. Reducing prosthesis mass is paramount \cite{Lenzi2018}, especially because mass attached to distal parts of the human body increases the metabolic energy consumed by the user. For example, in \cite{Waters1999}, a \SI{2}{\kilo\gram} load placed on each foot increased the rate of oxygen uptake 30\%, which is an indirect measure of metabolic energy consumption. SEAs have the potential to reduce mass in two ways: 1) extending battery life by reducing the energy dissipated by the actuator will allow the use of smaller batteries, and 2) reducing the speed-torque requirements for the motor will allow the use of smaller, lighter motors. As discussed in \cite{Bolivar2017}, an elastic element connected in series is passive and cannot reduce the energy required by the load, but it has the potential to reduce the energy dissipated for a given task. Reduction of dissipated energy is important, especially in tasks that are mainly dissipative such as level-ground walking. For example, the ankle of a \SI{75}{\kilo\gram} human provides about \SI{17}{\joule} per stride during normal walking, but a rigidly actuated prosthetic ankle consumes \SI{33}{\joule} \cite{Bolivar2017}. The same motor connected in series with an elastic element requires about \SI{25}{\joule} per stride, i.e., a 50\% reduction in the energy dissipated \cite{Bolivar2017}. Thus, SEAs represent a viable actuator alternative for the design of powered prosthetic legs, with some examples reported in \cite{Au2008,Rouse2014,Azocar2018}.



The explanation of our formulation starts in Section \ref{sec:ModelingOfSEA} with an introduction to the modeling of SEAs with an emphasis on their energy consumption. Section \ref{sec:EnergyConsumption} describes energy consumption as a convex-quadratic function of compliance. This quadratic expression is the cost function of our formulation. Section \ref{sec:RobustStiffness} describes our robust design methodology along with the actuator constraints. This section illustrates how to reformulate the optimization problem in order to find a solution that satisfies the constraints despite uncertainty. Section \ref{sec:CaseStudy} applies our methodology to the robust feasible design of SEAs for a powered prosthetic ankle. Section \ref{sec:DiscussionAndConclusion} includes the discussion and conclusion of our work.

\section{Modeling of Series Elastic Actuators} \label{sec:ModelingOfSEA}
\begin{figure}
	\centering
 	\includegraphics[scale = 1]{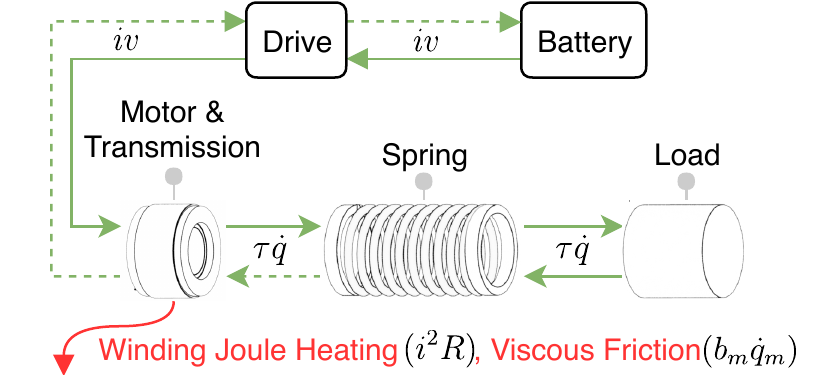}
    \caption{Energy flow of an SEA. Dashed lines indicate that the energy path may or may not exist depending on the construction of the device. For instance, energy flowing from the load to the battery requires that the load is high enough to backdrive the motor-transmission system.}
    \label{fig:EnergyFlowSEA}
\end{figure}

In this work, we assume that the SEA uses an electric DC motor, and energy can flow from the load to the energy source and vice versa. In other words, suitable electronics allow energy to flow to and from the battery. The corresponding energy flow and main components of an SEA are illustrated in Fig.~\ref{fig:EnergyFlowSEA}. We also assume that the energy consumption of the SEA refers to the energy consumption of the motor and energy lost at the transmission. The energy dissipated by viscous friction at the transmission can be lumped with the viscous friction of the motor's shaft. Energy losses in the power electronics and battery exist in practice, but we do not include them in our formulation. This is because they are either proportional to the energy losses of the motor (and hence can be lumped with winding losses) or are independent of the system's dynamics and hence are not affected by the spring stiffness. In addition, most of the losses occur at the motor's winding and transmission. For example, in the MIT Cheetah robot \cite{Seok2015}, 76\% of the total energy consumption is attributed to heat loss from the motor; the remaining energy is dissipated by friction losses and impacts \cite{Seok2015}. Under these assumptions, the energy consumption of the SEA is equivalent to the energy consumption of the motor including losses at the transmission, $E_m$, which is given by \cite{Rezazadeh2014}
\begin{equation}
E_m = \int_{t_0}^{t_f} \Big( \underbrace{\frac{\tau_m^2}{k_m^2}}_{\substack{\text{Winding} \\ \text{Joule} \\ \text{heating}}}+\underbrace{\tau_m\dot{q}_m }_{\substack{\text{Rotor} \\ \text{mechanical} \\ \text{power}}} \Big)  dt,
\label{eq:EnergyConsumedByMotor_IntegralExpression}
\end{equation}
where $t_0$ and $t_f$ are the initial and final times of the reference trajectory respectively, $k_m$ is the motor constant, $\tau_m$ the torque produced by the motor, and $\dot{q}_m$ the motor's angular velocity. Notice that energy associated with Joule heating can also be written as $i_m^2R$, since $\tau_m = i_m k_{t}$ and $k_m=k_{t}/\sqrt{R}$, where $i_m$ is the electric current flowing through the motor, $R$ the motor terminal resistance, and $k_{t}$ the motor torque constant \cite{Verstraten2015}.

Using the Newton-Euler method, the corresponding balance of torques at the motor and load side provides the following equations of motion \cite{spong:elastic,Verstraten2015}:
\begin{align}
I_m\ddot{q}_m &= -b_m\dot{q}_m+\tau_m+\frac{\tau_l}{\eta r}+\tau_u, \label{eq:EOM_SEA_MotorSide} \\ 
\tau_l &= g(q_l,\dot{q}_l,\ddot{q}_l,\tau_{\text{ext}}), \label{eq:EOM_SEA_LoadSide}
\end{align}
where $I_m$ is the inertia of the motor; $b_m$ its viscous friction coefficient; $r$ the transmission ratio; $\eta$ the efficiency of the transmission; $\tau_u$ is the unmodeled dynamics torque that lumps unmodeled effects at the motor and load side, e.g., cogging torque and friction; $q_l,\ \dot{q}_l,\ \ddot{q}_l,\ \tau_l$ represent the position, velocity, acceleration and torque of the load respectively; and $g(q_l,\dot{q}_l,\ddot{q}_l,\tau_{\text{ext}})$ defines the torque of the load based on the load dynamics and external torque, $\tau_{\text{ext}}$. For instance, in the case of an inertial load with viscous friction and an external torque, the load dynamics are defined by $g(q_l,\dot{q}_l,\ddot{q}_l,\tau_{\text{ext}}) = -I_l\ddot{q}_l-b_l\dot{q}_l+\tau_{\text{ext}}$, where $I_l$ is the inertia of the load, and $b_l$ its corresponding viscous friction coefficient. Because of the connection in series, the torque of the spring, $\tau_{s}$, is equal to the torque of the load, $\tau_{s} = \tau_{l}$. For a linear spring, the torque of the spring is proportional to its elongation, $\tau_s=k\delta$, where elongation is defined as
\begin{align}
\delta &= q_l-\frac{q_m}{r}.
\label{eq:ElongationDefinition}
\end{align}
 As seen in (\ref{eq:EOM_SEA_MotorSide})-(\ref{eq:EOM_SEA_LoadSide}), the elastic element cannot modify the torque required to perform the motion, $\tau_{\text{s}}$, but it can modify the position of the motor such that $I_m\ddot{q}_m+b_m\dot{q}_m$ reduces the torque of the motor, $\tau_m$.
\section{Energy Consumption as a Convex-Quadratic Function of Compliance} \label{sec:EnergyConsumption}

In the case of a linear spring, elongation and torque are related by $\tau_s = k(q_l - q_m/r)$, where $k$ is the stiffness constant. Using this relationship, the position of the motor and corresponding time derivatives can be expressed as a function of the given load position and the load torque $\tau_l$ as follows: $q_m = ( q_l - \tau_l/k) r$, $\dot{q}_m = ( \dot{q}_l - \dot{\tau}_l/k) r$, and $\ddot{q}_m = ( \ddot{q}_l - \ddot{\tau}_l/k) r$. Replacing these expressions into (\ref{eq:EOM_SEA_MotorSide}) and defining compliance as the inverse of stiffness, $\alpha := 1/k$, the expression of motor torque can be written as an affine function of compliance as follows:
\begin{align}
\tau_m &= \gamma_1 \alpha+\gamma_2, \label{eq:Tau simplified} \\
\noalign{where}
\gamma_1 &= -\left( I_m\ddot{\tau}_lr + b_m\dot{\tau}_lr \right), \label{eq:Gamma1}\\
\gamma_2 &= I_m\ddot{q}_lr+b_m\dot{q}_lr-\frac{\tau_s}{\eta r} - \tau_u, \label{eq:Gamma2}
\end{align}
are known constants that depend on the reference trajectory. Using the definition of $\tau_m$ in (\ref{eq:Tau simplified}), assuming periodic motion, and neglecting the uncertain torque, $\tau_u = 0$, we write the expression of energy consumption of the motor as the following convex-quadratic function of compliance:
\begin{align}
E_m &= \int_{t_0}^{t_f}  \left( \frac{ \tau_m^2}{k_m^2}+ \tau_m\dot{q}_m \right)  dt, \nonumber\\
&= \int_{t_0}^{t_f} \left( \frac{ \tau_m^2}{k_m^2}+ b_m\dot{q}_m^2-\frac{\tau_s\dot{q}_m}{\eta r} \right) dt + \int_{t_0}^{t_f} I_m\dot{q}_m d\dot{q}_m, \nonumber \\
 &= a\alpha^2+b\alpha+c\label{eq:EnergyAsACuadraticExpression},
\end{align}
where
\begin{align}
a &= \int_{t_0}^{t_f} \left( \dfrac{\gamma_1^2}{k_m^2}+b_mr^2\dot{\tau}_s^2\right) dt, \nonumber \\
b &= \int_{t_0}^{t_f} \left( \dfrac{2\gamma_1\gamma_2}{k_m^2}-2b_mr^2\dot{q}_l\dot{\tau}_s\right) dt, \nonumber \\
c &= \int_{t_0}^{t_f} \left( \dfrac{\gamma_2^2}{k_m^2} + b_m\dot{q}_l^2r^2-\dfrac{\dot{q}_l\tau_s}{\eta}\right) dt. \nonumber
\end{align}
\subsection*{Properties of the Convex-Quadratic Function of Compliance:}
\begin{enumerate}
\item $d^2E_m/d\alpha^2 = 2a \geq 0$, which follows from the definition of $a$. Therefore (\ref{eq:EnergyAsACuadraticExpression}) is a convex function of compliance \cite[p. 71]{boyd:convex}. \label{sec:ConvexityOfCostFunction}
\item Parameter $c$ is the energy consumed by a rigid actuator performing the same task without an elastic element, i.e., $$\lim_{k\to\infty} E_m = c.$$
\item The optimal value of compliance that minimizes energy consumption for any periodic trajectory is $\alpha = -b/(2a)$, neglecting actuator constraints. This optimal value can be computed in polynomial time. Note that the integrals in the definition of $a$ and $b$ can be approximated with discrete-time summations.
\item The sign of $b$ determines if the reference trajectories and motor configuration will benefit from series elasticity in order to reduce energy consumption. The quadratic cost function \eqref{eq:EnergyAsACuadraticExpression} leads to two possible scenarios for the effect of compliance $\alpha$ on motor energy (Fig.~\ref{fig:quadratic_ax2}). In the first case, $dE_m/d\alpha$ is negative at $\alpha=0$, thus series elasticity improves actuator efficiency over some range of compliance. In the second case, this slope is positive at $\alpha=0$, so energy increases with compliance, i.e., there is no energetic benefit to linear series elasticity for the given task. Thus, \textit{the necessary condition for an SEA to be energetically beneficial is $b<0$} in \eqref{eq:EnergyAsACuadraticExpression}, i.e.,
\begin{equation}
\mathlarger{\int}_{t_0}^{t_f} \left(\tfrac{2\gamma_1\gamma_2}{k_m^2}-2b_mr^2\dot{q}_l\dot{\tau}_s\right)dt<0.
\label{eq:NecessaryCondLinear}
\end{equation}
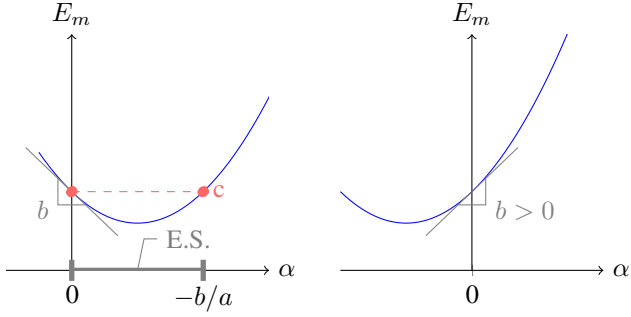
\begin{figure}
\centering
\begin{tikzpicture}[baseline]
\begin{axis}[
	width = 0.28\textwidth,
	height = 0.20\textheight,
	axis lines = middle,
	axis line style={->},
	xtick=\empty,ytick=\empty,
	xlabel= $\alpha$,
	ylabel= $E_m$,
	xlabel style={right},
	ylabel style={above},		
	xmin = -1,
	xmax = 3,
	ymin = 45,
	ymax = 60,
	xtick = {0.01, 2},
    xticklabels={0, $-b/a$},
]
	\newcommand\aP{2}
	\newcommand\bP{-4}
	\newcommand\cP{50}
	\addplot [
		blue,
		samples=201,
		domain =-0.5:4,
	]
		{\aP*x^2+\bP*x+\cP}
		coordinate [pos=0.31] (C)
	;
	\addplot [
		gray,
		solid,
		domain =-0.7:0.7,		
		samples = 200,
	]
		{\cP+\bP*x}
		coordinate [pos=0.35] (A)
		coordinate [pos=0.65] (B)
	;		
	\draw [gray](B) -| (A) 
		node [pos=0.75, anchor=north east]
		{$b$}
	;
	\addplot[
		red!60,
		dashed,
		mark = *,		
	]
		coordinates {(0,50) (2,50)}
	;
	\node [
		anchor = west, 
		red!60,
	] 
		at (C) {c}
	;
	\addplot[
		gray,
		solid,
		mark = |,	
		line width = 2,
		mark options = {
			scale = 2,
		},	
	]
		coordinates {(0,45) (2,45)}
	;
	\addplot[
		gray,
		solid,
		mark = none,		
	]
		coordinates {(1,45) (1.1,47) (1.3,47)}
		coordinate [pos=1] (Text)
	;	
	\node [
		anchor = west, 
		gray,
	] 
		at (Text) {E.S.}
	;
	\end{axis}
\end{tikzpicture}
    \hskip 10pt 
\begin{tikzpicture}[baseline]
\begin{axis}[
	width = 0.28\textwidth,
	height = 0.20\textheight,
	axis lines = middle,
	axis line style={->},
	xtick=\empty,ytick=\empty,
	xlabel= $\alpha$,
	ylabel= $E_m$,
	xlabel style={right},
	ylabel style={above},		
	xmin = -2,
	xmax = 2,
	ymin = 45,
	ymax = 60,
    xtick = {0.01},
    xticklabels={0},
]
	\newcommand\aP{2}
	\newcommand\bP{4}
	\newcommand\cP{50}
	\addplot [
		blue,
		samples=201,
		domain =-2:3,
	]
		{\aP*x^2+\bP*x+\cP}
		coordinate [pos=0.31] (C)
	;
	\addplot [
		gray,
		solid,
		domain =-0.7:0.7,		
		samples = 200,
	]
		{\cP+\bP*x}
		coordinate [pos=0.35] (A)
		coordinate [pos=0.65] (B)
	;		
	\draw [gray](A) -| (B) 
		node [pos=0.75, anchor=north west]
		{$b>0$}
	;
	\end{axis}
\end{tikzpicture}
	\caption{Left:~Energy consumption as a function of compliance, $\alpha$, where the energy savings (E.S.) region $0\leq \alpha \leq -b/a$ provides $E_m$ below the rigid level $c$. Right:~Case of motor and load that would not benefit energetically from series elasticity.}
    \label{fig:quadratic_ax2}
\end{figure}
\end{enumerate}

\section{Robust Stiffness Design} \label{sec:RobustStiffness}
This section presents our convex optimization formulation for the robust feasible design of the SEA's linear spring. 
The convex-quadratic function in (\ref{eq:EnergyAsACuadraticExpression}) is the cost function of our optimization problem. The constraints in our formulation are: 1) peak motor torque, 2) peak motor velocity, 3) limitations due to speed-torque relationship of DC motors, and 4) peak elongation of the spring. Below we discuss the definition, convexity, and uncertainty of the constraints.
\subsection{Actuator Constraints}
\subsubsection{Elongation Constraint}
Limited elongation of the elastic element is typical in SEA applications. An elastic element reaching its maximum elongation could be dangerous for co-robots. When the spring bottoms out, the elastic collisions with the environment become inelastic which may be harmful for the user and the robot itself. We express the elongation constraint as $\left\Vert \bm{\tau}_s \alpha \right\Vert_\infty \leq \delta_s$, where $\delta_s$ is the maximum elongation of the spring. This results in the constraint 
\begin{align}
\left\Vert m(\bm{\tau}_{l/m}) \alpha \right\Vert_\infty&\leq \delta_s, \nonumber \\
\begin{bmatrix}
m(\bm{\tau}_{l/m}) \\
-m(\bm{\tau}_{l/m})
\end{bmatrix}^T \alpha &\leq \bm{1} \delta_s, \nonumber\\
\bm{d}_1 \alpha &\leq \bm{e}_1, \label{eq:elongationConstraint}
\end{align}
where
\begin{equation}
\bm{d}_1 = 
\begin{bmatrix}
\bm{\tau}_{l/m} \\
-\bm{\tau}_{l/m}
\end{bmatrix},\ \bm{e}_1 =  \bm{1}\frac{\delta_s}{m}, \nonumber
\end{equation}
and $\bm{\tau}_{l/m}$ is a normalized expression of the load torque per unit of $m$, i.e., $\bm{\tau}_l = m \bm{\tau}_{l/m}$.
\subsubsection{Torque Constraint}\label{sec:TorqueConstraint}
In our formulation, we express the limitations in peak torque of the motor as $\left\Vert \bm{\tau}_m \right\Vert_\infty \leq \tau_{max}$, where $\tau_{max}$ is the maximum peak value of torque. Recall that the torque of the motor can be written as an affine function of compliance (\ref{eq:Tau simplified}), $\bm{\tau}_m = \bm{\gamma}_1\alpha+\bm{\gamma}_2$. Thus, constraining the peak torque is equivalent to the following affine constraint: 
\begin{align}
\left\Vert \bm{\gamma}_1\alpha + \bm{\gamma}_2 \right\Vert_\infty &\leq \tau_{max}, \nonumber \\
\begin{bmatrix}
\bm{\gamma}_1 \\
-\bm{\gamma}_1
\end{bmatrix} \alpha  &\leq \bm{1} \tau_{max} + \begin{bmatrix}
-\bm{\gamma}_2 \\
\bm{\gamma}_2
\end{bmatrix}, \nonumber \\
\bm{d}_2\alpha &\leq \bm{e}_2, \label{eq:TorqueConstraint}
\end{align}
where
\begin{align*}
\bm{d}_2 &= \begin{bmatrix}
I_m\bm{\ddot{\tau}}_{l/m}r + b_m\bm{\dot{\tau}}_{l/m}r \\
-I_m\bm{\ddot{\tau}}_{l/m}r - b_m\bm{\dot{\tau}}_{l/m}r 
\end{bmatrix}, \nonumber \\[0.2cm]
\bm{e}_2 &= \begin{bmatrix}
\dfrac{\bm{\tau}_{l/m}}{\eta r} \\
\dfrac{-\bm{\tau}_{l/m}}{\eta r}
\end{bmatrix} 
 + \begin{bmatrix}
-I_m r\bm{\ddot{q}}_l - b_m r\bm{\dot{q}}_l + \bm{1} (\tau_{max} + \tau_u)\\
I_m r \bm{\ddot{q}}_l + b_m r\bm{\dot{q}}_l + \bm{1} (\tau_{max} - \tau_u)
\end{bmatrix}\frac{1}{m}. \nonumber
\end{align*}
\subsubsection{Speed-Torque Relationship Constraint}
As an actuator, a DC motor simultaneously operates as an electric generator producing a back-emf voltage. This back-emf voltage, which is proportional to the motor's speed of rotation, limits the current that can flow through the motor's winding, which is proportional to the torque produced by the motor. As a consequence, the torque that a DC motor generates is a function of the rotational speed \cite[p. 536]{carryer2011}. This phenomenon is summarized by the equation $\tau_m(R/k_t) = v_{in} - k_t\dot{q}_m$, where $v_{in}$ is the input voltage to the electric motor. Then for a DC motor to be feasible $\tau_m(R/k_t) \leq v_{in} - k_t\dot{q}_m$ \cite{Rezazadeh2014}. The same inequality applies for positive and negative values of speed and torque, therefore in total there are four inequalities to express the torque-velocity relationship constraints. The following affine constraint represents these inequalities: 
\begin{align}
\bm{\tau}_m &\leq \bm{1}v_{in}\frac{k_t}{R} - \frac{k_t^2}{R}\bm{\dot{q}}_m, \nonumber \\
\bm{\gamma}_1 \alpha +\bm{\gamma}_2 &\leq \bm{1}v_{in} \frac{k_t}{R} - \frac{k_t^2}{R}(\bm{\dot{q}}_l-\bm{\dot{\tau}}_l\alpha)r, \nonumber \\
\bm{d}_{3a}\alpha &\leq \bm{e}_{3a}, \label{eq:TorqueSpeedConstraint}
\end{align}
where
\begin{align}
\bm{d}_{3a} &= I_m\bm{\ddot{\tau}}_{l/m}r + b_m\bm{\dot{\tau}}_{l/m}r - \frac{k_t^2r}{R} \bm{\dot{\tau}}_{l/m}, \nonumber \\
\bm{e}_{3a} &= \frac{\bm{\tau}_{l/m}}{\eta r}+\left(\bm{1} \left( v_{in} \frac{k_t}{R}+\tau_u \right)-I_mr\bm{\ddot{q}}_l- \right. \nonumber \\
& \left. b_mr\bm{\dot{q}}_l -\frac{k_t^2r}{R} \bm{\dot{q}}_l \right) \frac{1}{m}. \nonumber
\end{align}
Using positive and negative values of torque and speed we can define three similar versions of the inequality (\ref{eq:TorqueSpeedConstraint}), which we will denote using the vectors $\bm{d}_{3_{b, c, d}}$ and $\bm{e}_{3_{b, c, d}}$. Summarizing, the torque and velocity relationship constraints can be lumped into the single vector inequality constraint
\begin{equation}
\bm{d}_3 \alpha \leq \bm{e}_3, \label{eq:TorqueSpeedConstraintLumped}
\end{equation}
where
\begin{align}
\bm{d}_3 = [\bm{d}_{3a}^T, \bm{d}_{3b}^T, \bm{d}_{3c}^T, \bm{d}_{3d}^T]^T,\ \bm{e}_3 = [\bm{e}_{3a}^T, \bm{e}_{3b}^T, \bm{e}_{3c}^T, \bm{e}_{3d}^T]^T. \nonumber
\end{align}
\subsubsection{RMS Torque and Maximum Speed}
Long-term operation of an electric motor can generate excessive heat and can be harmful for the actuator. Constraining the RMS torque is a typical method to guarantee that long-term operation is safe for the device. In our formulation, the square of the RMS torque can be written as a convex-quadratic function of compliance, and therefore can be included as a constraint. However, RMS torque also appears in our cost function (\ref{eq:EnergyAsACuadraticExpression}). Therefore, it is redundant to include it as a constraint. The constraint (\ref{eq:TorqueSpeedConstraintLumped}) already considers the maximum speed of rotation of the motor, which is equivalent to $\tau_m(R/k_t) \leq v_{in} - k_t\dot{q}_m$ when the motor torque, $\tau_{m}$, is zero.
\subsubsection{Lumping the Constraints}
Peak motor torque, peak motor velocity, speed-torque relationship constraints, and maximum elongation of the spring can be represented as the following vector inequalities:
\begin{align}
\bm{d}\alpha &\leq \bm{e} \label{eq:constraintsLumped}
\end{align}
where
\begin{align}
\bm{d} = [\bm{d}_1^T, \bm{d}_2^T, \bm{d}_3^T]^T,\ \bm{e} = [\bm{e}_1^T, \bm{e}_2^T, \bm{e}_3^T]^T.
\end{align}

\subsection{System Uncertainty} \label{subsubsec:Uncertainty}
Feasibility of the constraints is subject to the selection of the spring compliance and uncertainty in the definition of the constraints. Uncertainty in our formulation means that the reference kinematics and kinetics of the load, the manufacturing accuracy of the spring, the efficiency of the transmission, and the unmodeled dynamics are not precisely known but are restricted to belong to an uncertainty set, $\mathcal{U}$. In our formulation, $\mathcal{U}$ is defined as the Cartesian product
\begin{equation*}
\mathcal{U} = \mathcal{U}_{q_l} \times \mathcal{U}_{\dot{q}_l} \times \mathcal{U}_{\ddot{q}_l} \times \mathcal{U}_{m} \times \mathcal{U}_{\eta} \times \mathcal{U}_{\tau_u} \times \mathcal{U}_{d},
\end{equation*}
where the uncertainty sets $\mathcal{U}_{q_l}, \mathcal{U}_{\dot{q}_l}, \mathcal{U}_{\ddot{q}_l}, \mathcal{U}_{m}, \mathcal{U}_{\eta}, \mathcal{U}_{\tau_u},\text{and}\ \mathcal{U}_{d}$ express all the possible realizations for the load position, velocity, and acceleration; the multiplicative factor of the load torque; the efficiency of the transmission; the unmodeled dynamics; and the manufacturing accuracy of the spring respectively.

For the position of the load, the uncertainty set is defined as follows:
\begin{equation}
\mathcal{U}_{q_l} = \{ \bm{q}_{l} \in \mathbf{R}^n: \bm{\bar{q}}_{l}-\bm{1}\varepsilon_{q_l}\leq \bm{q}_{l} \leq \bm{\bar{q}}_{l}+\bm{1}\varepsilon_{q_l} \}, \nonumber
\end{equation} 
where $\bm{\bar{q}}_l \in \mathbf{R}^n$ and $\varepsilon_{q_l} \in \mathbf{R}$ represent the nominal load trajectory and uncertainty of the load position, respectively. Inequalities for vectors are element-wise. In other words, the position of the load, $\bm{q}_l$, is within $\bm{\bar{q}}_l\pm \bm{1}\varepsilon_{q_l}$. Using the respective nominal and uncertainty values ($\bm{\dot{\bar{q}}}_l, \bm{\ddot{\bar{q}}}_l, \bar{\eta}, \bar{\tau}_u, \varepsilon_{\dot{q}_l}, \varepsilon_{\ddot{q}_l}, \varepsilon_{\eta}, \varepsilon_{\tau_u}$), we use the same definition for $\mathcal{U}_{\dot{q}_l}, \mathcal{U}_{\ddot{q}_l}, \mathcal{U}_{\eta},\text{and}\ \mathcal{U}_{\tau_u}$. Uncertainty in the manufacturing of the spring is defined as the factor $(1\pm\varepsilon_d)$ that multiplies the spring compliance. Because it is a multiplicative factor, uncertainty in the manufacturing of the spring is equivalent to uncertainty in the coefficient vector $\bm{d}$, as seen in (\ref{eq:constraintsLumped}). Therefore the corresponding uncertainty set is defined by $$\mathcal{U}_{d} = \{\bm{d} \in \mathbf{R}^p: \bm{d}-\varepsilon_d|\bm{d}| \leq \bm{d} \leq \bm{d}+\varepsilon_d|\bm{d}|\},$$ where $p$ is the number of constraints. Inequalities and absolute values for vectors are element-wise. This uncertainty in the manufacturing of the spring implies that its stiffness is in the set $$k \in \{k \in \mathbf{R}: [(1+\varepsilon_d)\alpha]^{-1} \leq k \leq [(1-\varepsilon_d)\alpha]^{-1} \}.$$ Uncertainty in the kinetic reference trajectories is defined by a nominal value and a uncertain multiplicative factor. Precisely, the reference torque of the load $\bm{\tau}_l$ is considered to be $\bm{\tau}_l~=~m(\bm{\tau}_{l/m})$, where $\bm{\tau}_{l/m}$ is a nominal value of $\bm{\tau}_l$ per unit of $m$. Our uncertain multiplicative factor, $m$, could be any element within the set 
\begin{equation}
\mathcal{U}_{m} = \{ m \in \mathbf{R}: 0< \bar{m}-\varepsilon_{m} \leq m \leq \bar{m}+\varepsilon_{m}\}, \nonumber
\end{equation} 
where $\bar{m} \in \mathbf{R}$ is the nominal value of $m$ and $\varepsilon_{m} \in \mathbf{R}$ is its corresponding uncertainty. In other words, $\bm{\tau}_l = (\bar{m} \pm \varepsilon_{m}) \bm{\tau}_{l/m}$.

\subsection{The Robust Formulation of the Constraints}
A robust feasible design should satisfy the constraints (\ref{eq:constraintsLumped}) for all possible realizations of the uncertainty within the uncertainty set. Note that the uncertainty in the manufacturing of the spring manifests as uncertainty in the vector $\bm{d}$ in  (\ref{eq:constraintsLumped}). Thus, a robust feasible design results in an optimal selection of $\alpha$ that satisfies
\begin{equation}
\bm{d} \alpha \leq \bm{e},\ \forall \bm{q}_l, \bm{\dot{q}}_l, \bm{\ddot{q}}_l, m, \eta, \tau_u, \bm{d} \in \mathcal{U}.
\end{equation}
Because $\alpha > 0$, a robust feasible solution is equivalent to
\begin{equation}
\bar{\bm{d}} \alpha \leq \underline{\bm{e}}, \label{eq:robustConstraints}
\end{equation}
where $\bar{\bm{d}}$ and $\underline{\bm{e}}$ are the vectors that represent the worst case representation of the uncertainty. These vectors are defined as follows:
\begin{equation}
\bar{\bm{d}} = \bm{d} + \varepsilon_{d}|\bm{d}|,\ \underline{\bm{e}} = [\underline{\bm{e}}_1^T, \underline{\bm{e}}_2^T, \underline{\bm{e}}_3^T]^T,
\end{equation}
where
\begin{align*}
\underline{\bm{e}}_1 &=  \bm{1}\frac{\delta_s}{\bar{m}+\varepsilon_m}, \\
\underline{\bm{e}}_2 &= \begin{bmatrix}
\bm{\tau}_{l/m} \\
-\bm{\tau}_{l/m}
\end{bmatrix} \frac{1}{(\bar{\eta}\pm\varepsilon_\eta)} r
 + \bm{f} \frac{1}{\bar{m}\pm\varepsilon_m}, \\[0.2cm]
\bm{f} &=\begin{bmatrix}
-I_m r(\bm{\ddot{q}}_l+ \bm{\varepsilon}_{\ddot{q}_l}) - b_m r(\bm{\dot{q}}_l+ \bm{\varepsilon}_{\dot{q}_l}) + \bm{1}(\underline{\tau_{u}}+\tau_{max})\\
I_m r(\bm{\ddot{q}}_l- \bm{\varepsilon}_{\ddot{q}_l}) + b_m r(\bm{\dot{q}}_l-\bm{\varepsilon}_{\dot{q}_l}) + \bm{1}(\underline{\tau_{u}}+\tau_{max})
\end{bmatrix}, \\[0.2cm]
\underline{\bm{e}}_3 &= [\underline{\bm{e}}_{3a}^T, \underline{\bm{e}}_{3b}^T, \underline{\bm{e}}_{3c}^T, \underline{\bm{e}}_{3d}^T]^T, \\
\underline{\bm{e}}_{3a} &= \frac{\bm{\tau}_{l/m}}{(\bar{\eta}\pm\varepsilon_\eta) r}+(-I_m(\bm{\ddot{q}}_l+\varepsilon_{\ddot{q}_l})r-b_m(\bm{\dot{q}}_l+\varepsilon_{\dot{q}_l})r + \nonumber\\
& \bm{1} \left( v_{in} \frac{k_t}{R}+\underline{\tau_u}\right) - \frac{k_t^2r}{R} (\bm{\dot{q}}_l+\varepsilon_{\dot{q}_l}))\frac{1}{(\bar{m}\pm\varepsilon_{m})},
\end{align*}
and the values for $\underline{\bm{e}}_{3b}, \underline{\bm{e}}_{3c}$ , and $\underline{\bm{e}}_{3d}$ are defined using positive and negative values of torque and speed in the definition of the torque-speed constraints. The sign of ${1}/{(\bar{m}\pm\varepsilon_{m})}$ depends on the elements of the vector that it multiplies, as the multiplication applies element-wise. When the element of the vector is positive, then the multiplication factor becomes ${1}/{(\bar{m}+\varepsilon_{m})}$; when the element is negative, ${1}/{(\bar{m}-\varepsilon_{m})}$. The same idea applies to ${1}/{(\bar{m}\pm\varepsilon_{\eta})}$, which describes the worst possible scenario to satisfy the inequality (\ref{eq:robustConstraints}).

\subsection{The Optimization Problem}
Combining the definition of the cost function in (\ref{eq:EnergyAsACuadraticExpression}), and the constraints in (\ref{eq:robustConstraints}), the optimization problem becomes
\begin{equation}
\begin{aligned}
& \underset{\alpha}{\text{minimize}}
& & a\alpha^2+b\alpha+c,\\
& \text{subject to} & & \bar{\bm{d}}\alpha \leq \underline{\bm{e}},
\end{aligned} \label{eq:THEoptimizationProblem}
\end{equation}
also known as a convex-quadratic program with affine inequality constraints. This is a convex-optimization problem; the cost function is convex as shown in Section \ref{sec:ConvexityOfCostFunction} and the constraints represent a closed interval which is a convex set described by the affine inequality (\ref{eq:robustConstraints}). The solution of this optimization problem is a value of compliance that is robust feasible, i.e., it satisfies the actuator constraints despite uncertainty in the load.

\section{Case Study: Simulation of a Powered Prosthetic Ankle}\label{sec:CaseStudy}

\begin{figure}
	\centering
	\def\svgscale{1.5}
	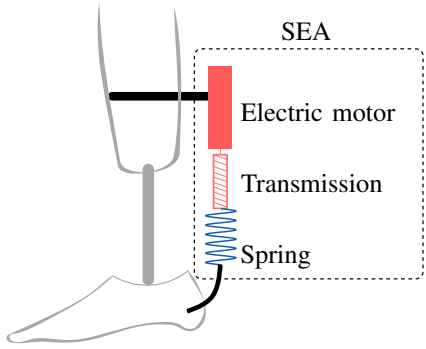
	\caption{Schematic SEA for powered prosthetic ankle.}
	\label{fig:schematicSEA}
\end{figure}

In this section, we apply our methods to the design of an SEA for a powered prosthetic ankle to minimize energy consumption while satisfying actuator constraints despite uncertainty. Figure \ref{fig:schematicSEA} illustrates a schematic of the prosthesis. Traditionally, actuator designs for powered prostheses use average kinetic and kinematic trajectories \cite{Rouse2014, Au2008, Lenzi2018, Elery2018}. However, load conditions during human locomotion vary significantly even for a single subject \cite{Embry2016:EMBC}. Robust design is important in this application as human locomotion and manufacturing methods are inherently uncertain. For instance, the ankle joint position during human locomotion varies with a standard deviation of $\pm$\SI{5}{\degree} \cite{Winter1983}, and the stiffness of a manufactured spring has a standard deviation of about $\pm$10\% from the desired stiffness value \cite{Azocar2018}.  

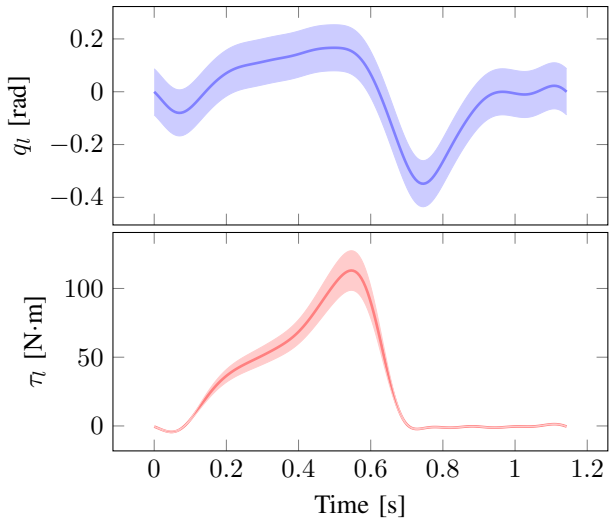
\begin{figure}
\centering
\setlength{\abovecaptionskip}{5pt}
\begin{tikzpicture}
	\begin{groupplot}[
		group style={
			group size=1 by 2,
			xlabels at = edge bottom,	
			vertical sep = 0.1cm,		
			},
		width = 0.45\textwidth, 
		height = 0.19\textheight,		
		xlabel = Time  \lbrack\SI{}{\second}\rbrack,		
		]
	\nextgroupplot [
		ylabel  = $q_l$ \lbrack\SI{}{\radian}\rbrack,
		xtick = {},
		xticklabels = \empty,
		]
	\addplot [blue!50, yticklabel pos=right,line width=1pt] table[x=time,y=ql]{UncertainReferenceTrajectory.dat};
    \addplot [blue!20, solid, line width=0pt, name path=A,] table[x=time,y=qlU]{UncertainReferenceTrajectory.dat};
    \addplot [blue!20, solid, line width=0pt, name path=B,] table[x=time,y=qlL]{UncertainReferenceTrajectory.dat};
    \addplot [blue!20] fill between[of=A and B,];
	\nextgroupplot [
		ylabel  = $\tau_l$ \lbrack\SI{}{\newton\meter}\rbrack,
		]
	\addplot [red!50, yticklabel pos=right,line width=1pt] table[x=time,y=tau]{UncertainReferenceTrajectory.dat};
    \addplot [red!20, solid, line width=0pt, name path=A,] table[x=time,y=tauU]{UncertainReferenceTrajectory.dat};
    \addplot [red!20, solid, line width=0pt, name path=B,] table[x=time,y=tauL]{UncertainReferenceTrajectory.dat};
    \addplot [red!20] fill between[of=A and B,];	
	\end{groupplot}
\end{tikzpicture}
\caption{Position (top) and torque (bottom) of the human ankle during level ground walking \cite{Winter1983}. The solid line indicates the mean values for a \SI{69.1}{\kilo\gram} subject \cite{Winter1983}. The shaded region around the nominal trajectory illustrates the uncertainty in the position $\varepsilon_{q_l} = \pm \SI{5}{\degree}$ and the mass of the subject $\varepsilon_{m} = \pm  \SI{8.8}{\kilo\gram}$. This uncertainty corresponds to the standard deviation reported in \cite{Winter1983}.}
\label{fig:referenceUncertainTrajectories}
\end{figure}

In our formulation, we take advantage of the connection between uncertainty in the kinematics and kinetics of the load and our definition of uncertainty sets to obtain a robust feasible design. The parameters $\varepsilon_{q_l}, \varepsilon_{\dot{q}_l}, \varepsilon_{\ddot{q}_l}$ define the uncertainty in the kinematics $\mathcal{U}_{\{q_l, \dot{q}_l, \ddot{q}_l\}}$. In this simulation case study, we define these parameters to be equal to the reported standard deviation of the joint kinematics in \cite{Winter1983}. Our formulation of uncertainty in the kinetics has a practical meaning in biomechanics. The reference torque of the ankle joint is traditionally normalized by the mass of the user \cite{winter2009biomechanics}. Because our definition of uncertainty in the kinetics is multiplicative, it is equivalent to uncertainty in the user's mass. As a result, it becomes relevant to rehabilitation and physical assistance robots where users can vary or a single user can wear additional accessories, such as backpacks. We select the uncertainty in the mass, $\varepsilon_m$, to be equal to the reported standard deviation of the subjects' mass in \cite{Winter1983}. Figure \ref{fig:referenceUncertainTrajectories} illustrates the reference trajectories and corresponding bounds of uncertainty. Uncertainty in the manufacturing of the spring, $\varepsilon_d$, is equal to twice the standard deviation of the SEA spring stiffness of the open-source prosthetic leg at University of Michigan \cite{Azocar2018}. The uncertain torque, $\varepsilon_{\tau_u}$, is equal to \SI{10}{\percent} of the maximum continuous motor torque. Uncertainty in the efficiency of the transmission is based on our experience aiming for a realistic simulation case. Table \ref{table:RefParameters} illustrates the parameters of the actuator and Table \ref{table:uncertainParameters} the parameters of uncertainty. The parameters of the actuator are inspired by the design of the first-generation powered prosthetic leg at the University of Texas at Dallas \cite{Quintero2016:IROS, Quintero2018}. Using the actuator parameters and reference trajectories, we used CVX, a package for specifying and solving convex programs \cite{cvx,Grant2008}, to solve the optimization problem (\ref{eq:THEoptimizationProblem}).

\begin{table}
\renewcommand{\arraystretch}{1.3}
\caption{Motor Parameters (EC-30 from Maxon motor).}
\label{table:RefParameters}
\centering
\begin{tabular}{lcc}
\hline
Parameter	&EC30 & Units\\
\hline
Motor torque constant, $k_t$ & 13.6 & \SI{}{\milli\newton\meter\per\ampere}\\
Motor terminal resistance, $R$ & 102 & \SI{}{\milli\ohm}\\
Motor inertia, $I_m$ & 33.3 & \SI{}{\gram\square\centi\meter}\\
Gear ratio, $r$ & 600 & \\
Efficiency transmission, $\eta$ & 0.8 & \\
Motor viscous friction, $b_m$ & 1.665 & \SI{}{\micro\newton\meter\second\per\radian}\\
Max. motor torque, $\tau_{max}$ & 337.5 & \SI{}{\milli\newton\meter}\\
Max. motor velocity, $\dot{q}_{max}$ & 21065 & \SI{}{rpm}\\
Voltage, $v_{in}$ & 30 & \SI{}{\volt}\\
\hline
\end{tabular}
\end{table}

\begin{table}
\renewcommand{\arraystretch}{1.3}
\caption{Uncertainty based on the variance in \cite{Winter1983,Azocar2018}.}
\label{table:uncertainParameters}
\centering
\begin{tabular}{ll}
\hline
Uncertainty in	&Units\\
\hline
Mass, $\varepsilon_m$ & $\pm$\SI{8.8}{\kilo\gram} \\
Reference position, $\varepsilon_{q_l}$ & $\pm$\SI{5}{\degree} \\
Reference velocity, $\varepsilon_{\dot{q}_l}$ & $\pm$\SI{30}{\percent} rms average trajectory\\
Reference acceleration, $\varepsilon_{\ddot{q}_l}$ & $\pm$\SI{30}{\percent} rms average trajectory\\
Transmission efficiency, $\varepsilon_{\eta}$ & $\pm$\SI{20}{\percent} \\
Unmodeled dynamics, $\varepsilon_{\tau_u}$ & $\pm$\SI{13.5}{\milli\newton\meter}\\
Manufacturing of spring, $\varepsilon_{d}$ & $\pm$\SI{20}{\percent} \\
\hline
\end{tabular}
\end{table}

\subsection*{Results}

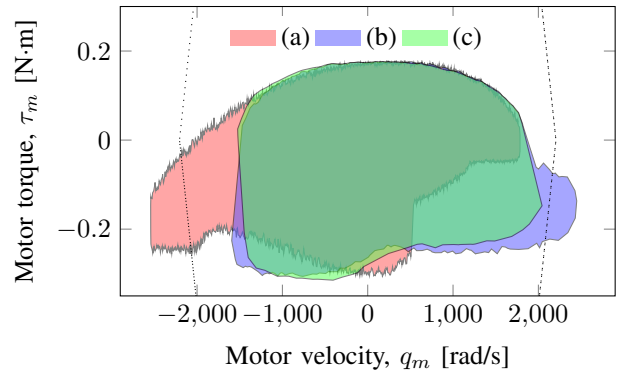
\begin{figure}
\centering
\begin{tikzpicture}
\begin{axis}[ 
    xlabel = {Motor velocity, $q_m$ [\SI{}{\radian\per\second}]},
    ylabel = {Motor torque, $\tau_m$ [\SI{}{\newton\meter}]},
    width = 0.45\textwidth, 
	height = 0.23\textheight,
    ymin = -0.35,
    ymax = 0.3,
    xmin = -2900,
    xmax = 2900,
    legend image code/.code={%
                    \draw[#1, draw=none] (0cm,-0.1cm) rectangle (0.6cm,0.1cm);
                },
    legend style = {draw=none, 
    				legend columns=-1,
                    at={(axis cs:1500,0.28), anchor = north east},
                    },
	]
    \addplot+ [black, fill = red!70, opacity=0.5, no markers] table[x = rigidQmd, y=rigidTaum]{rigid_Actuator.dat};
	\addplot+ [black, fill = blue!70, opacity=0.5, no markers] table[x = nominalQmd, y=nominalTaum]{nominal_Actuator.dat};
    \addplot+ [black, opacity=0.5, fill = green!70, no markers] table[x = robustQmd, y=robustTaum]{robust_Actuator.dat};    
    \addplot [black, dotted] table[x = speed, y = torque, forget plot]{motorSpeedTorqueLim.dat}; 
    \legend{(a), (b), (c)}
\end{axis}
\end{tikzpicture}
\caption{Speed and torque requirements of different actuators for a powered prosthetic ankle. The region enclosed by the dotted line describes the speeds and torques that satisfy the specifications of the motor, i.e., feasible region. Figure shows three possible actuator designs: (a) rigid actuator Maxon EC-30 without series elasticity, (b) SEA using the same motor with optimal stiffness that satisfies constraints only for the nominal data, and (c) SEA with the same motor that satisfies actuator constraints despite uncertainty using our formulation. The robust design (c) is the only actuator that satisfies the actuator constraints for all possible values of uncertainty.}
\label{fig:SpeedAndTorqueRequirements}
\end{figure}

To contextualize our results, we analyze three possible actuator designs: (a) a rigid actuator Maxon EC-30 without series elasticity, (b) an SEA using the same motor with optimal stiffness that satisfies constraints only for the nominal data, and (c) an SEA with the same motor that satisfies actuator constraints despite uncertainty using our robust formulation. Using (\ref{eq:EOM_SEA_MotorSide}) and (\ref{eq:EOM_SEA_LoadSide}) we compute the motor speed and torque trajectories considering the ankle kinematics and kinetics as the load. Figure \ref{fig:SpeedAndTorqueRequirements} illustrates these trajectories in a torque-speed plot. For the actuator (a), the required speed and torque do not stay within the specifications of the motor and therefore the rigid actuator is infeasible. Including series elasticity, the design (b) makes the actuator feasible and dissipates 30.8\% less energy compared to (a). This justifies the use of series elasticity, not only for the reduction of energy consumption, but also to maintain the requirements within the actuator specifications. The optimal stiffness of design (b) is \SI{217.4}{\newton\meter\per\radian}. However, this design becomes infeasible when the reference trajectory deviates within the uncertainty set, as shown in Fig. \ref{fig:SpeedAndTorqueRequirements}. Using our robust formulation, design (c) satisfies the constraints despite uncertainty using a spring stiffness of \SI{243.4}{\newton\meter\per\radian}. Design (c) reduces 30.45\% of the dissipated energy compared to a 30.8\% reduction in the case (b), where the reported energy savings are relative to the rigid case. The small trade-off in performance using the robust SEA is justified when feasibility of the constraints is satisfied. Table \ref{table:summaryResults} summarizes the results.
\begin{table}
\renewcommand{\arraystretch}{1.3}
\caption{Optimization results that indicate weak trade-off between robustness and energy savings. Energy savings are relative to dissipated energy of the rigid actuator \SI{11.7}{\joule}.}
\label{table:summaryResults}
\centering
\begin{tabular}{lcc}
\hline
Design	& Optimal Stiffness & Energy Savings\\
\hline
nominal (b) &  \SI{217.4}{\newton\meter\per\radian} & 30.8\% \\
robust feasible (c) & \SI{243.4}{\newton\meter\per\radian} & 30.45\% \\
\hline
\end{tabular}
\end{table}

\section{Discussion and Conclusion} \label{sec:DiscussionAndConclusion}

In this paper, we introduced a convex optimization formulation to compute the stiffness of SEAs that minimizes energy consumption and satisfies actuator constraints despite uncertainty due to manufacturing of the spring, unmodeled dynamics, efficiency of the transmission, and the kinematics and kinetics of the load. The methodology relies on the following two concepts: a scalar convex-quadratic function of compliance to express motor energy consumption, and defining uncertainty sets that represent tractable solutions of the optimization problem. As shown in our simulation case study, series elasticity can reduce energy consumption and also modify the speed and torque of the motor so that it becomes feasible. 

Our simulation case study illustrated the robust feasible SEA design for a powered prosthetic ankle. Uncertainty from the recorded biomechanics naturally connected with our definition of the uncertainty sets. The results illustrate that a small trade off between robustness and energy consumption justifies a robust feasible design. It is important to note that the robust solution satisfies actuator constraints despite the uncertainty described in Table \ref{table:uncertainParameters}. Previous research \cite{RobertBrown2013} did not consider a robust feasible solution of the optimization problem, however, they analyzed the effect of uncertainty in the energetic cost. Their results indicate that as the required motion of an SEA becomes more arbitrary, the optimal spring stiffness that minimizes power consumption approaches infinity, showing that the best design for a completely arbitrary task is a system without spring. In general, our results indicated a similar trend: the more arbitrary or the bigger the uncertainty sets, the stiffer the optimal design. However, when considering feasibility of the actuator, infinite spring stiffness may lead to an infeasible actuator. Thus, a robust feasible optimal solution cannot be obtained simply by increasing stiffness. Instead, it requires proper treatment of uncertainty as presented in our convex optimization method.


The convex-quadratic expression of compliance in (\ref{eq:EnergyAsACuadraticExpression}) is beneficial beyond our robust formulation. The convexity and simplicity of the expression allow optimization algorithms to find the optimal value of stiffness in polynomial time \cite{Nesterov94}. This could be exploited by VSAs to calculate their reference stiffness values during operation. In the unconstrained case, the proposed convex-quadratic expression has an analytical solution, which is useful to study the principles of series elasticity. For instance, (\ref{eq:NecessaryCondLinear}) describes the necessary conditions for periodic trajectories so that series elasticity can reduce energy consumption. Future work will focus on experimental applications and extend the presented robust formulation to the robust design of SEAs that use nonlinear springs. The design of nonlinear springs for SEAs can be formulated as a discrete-time convex optimization problem \cite{Bolivar2018}, which is desirable for a robust formulation.

\bibliographystyle{IEEEtran}
\bibliography{IEEEabrv,references2}

\end{document}